\documentclass[10pt,twocolumn,letterpaper]{article}

\usepackage{cvpr}              

\usepackage{graphicx}
\usepackage{amsmath}
\usepackage{amssymb}
\usepackage{booktabs}

\usepackage[pagebackref,breaklinks,colorlinks]{hyperref}

\usepackage[capitalize]{cleveref}
\crefname{section}{Sec.}{Secs.}
\Crefname{section}{Section}{Sections}
\Crefname{table}{Table}{Tables}
\crefname{table}{Tab.}{Tabs.}

\def\cvprPaperID{*****} 
\def\confName{CVPR}
\def\confYear{2022}

\begin{document}

\title{1st Place Solution for YouTubeVOS Challenge 2022: \\ Referring Video Object Segmentation}

\author{Zhiwei Hu\footnotemark[2], Bo Chen\footnotemark[2], Yuan Gao, Zhilong Ji, Jinfeng Bai\\
\small $^1$Tomorrow Advancing Life (TAL) Education Group \\
{\tt\small \{huzhiwei3, chenbo2, gaoyuan23, jizhilong, baijinfeng1\}@tal.com}
}
\maketitle
\thispagestyle{empty}
\renewcommand{\thefootnote}{\fnsymbol{footnote}}
\footnotetext[2]{Equal Contribution}

\begin{abstract}
   The task of referring video object segmentation aims to segment the object in the frames of a given video to which the referring expressions refer. Previous methods adopt multi-stage approach and design complex pipelines to obtain promising results. Recently, the end-to-end method based on Transformer has proved its superiority. In this work, we draw on the advantages of the above methods to provide a simple and effective 
  pipeline for RVOS.  Firstly, We improve the state-of-the-art one-stage method ReferFormer to obtain mask sequences that are strongly correlated with language descriptions. Secondly, based on a reliable and high-quality keyframe, we leverage the superior performance of video object segmentation model to further enhance the quality and temporal consistency of the mask results. Our single model reaches 70.3 $\mathcal{J} \& \mathcal{F}$ on the Referring Youtube-VOS validation set and 63.0 on the test set. After ensemble, we achieve 64.1 on the final leaderboard, ranking 1st place on CVPR2022 Referring Youtube-VOS challenge. Code will be available at \url{https://github.com/Zhiweihhh/cvpr2022-rvos-challenge.git}.
\end{abstract}

\section{Introduction}
\label{sec:intro}

Referring video object segmentation(RVOS) is a task of segmenting the target instance in the frames of a given video based on the natural language expression. Compared with traditional video object segmentation, RVOS requires understanding both visual and textual content and locating the referred object based on cross-modal reasoning, which is a more challenging task. RVOS has more convenience in applications such as human-computer interaction and video editing, thus has received wide attention from the community.

To achieve better performance, existing methods usually adopt multi-stage approach and design complex pipelines, which suffer from poor scalability and optimization difficulties. For example, The previous champion\cite{liang2021rethinking} of this track proposed a three-stage approach, including an instance segmentation module, a propagation module and a tracklet-language grounding module to achieve encouraging results. However, this method requires separately tuning the performance of each module which may lead to sub-optimal solution. Recently, inspired by Transformer~\cite{vaswani2017attention} and DETR~\cite{carion2020end}, ReferFormer~\cite{wu2022referformer} proposed a simple end-to-end framework for the RVOS task. This method views the language as queries and directly attends to the most relevant regions in the video frames, resulting in state-of-the-art performance.

\begin{figure}[t]
\begin{center}
\includegraphics[width=85mm]{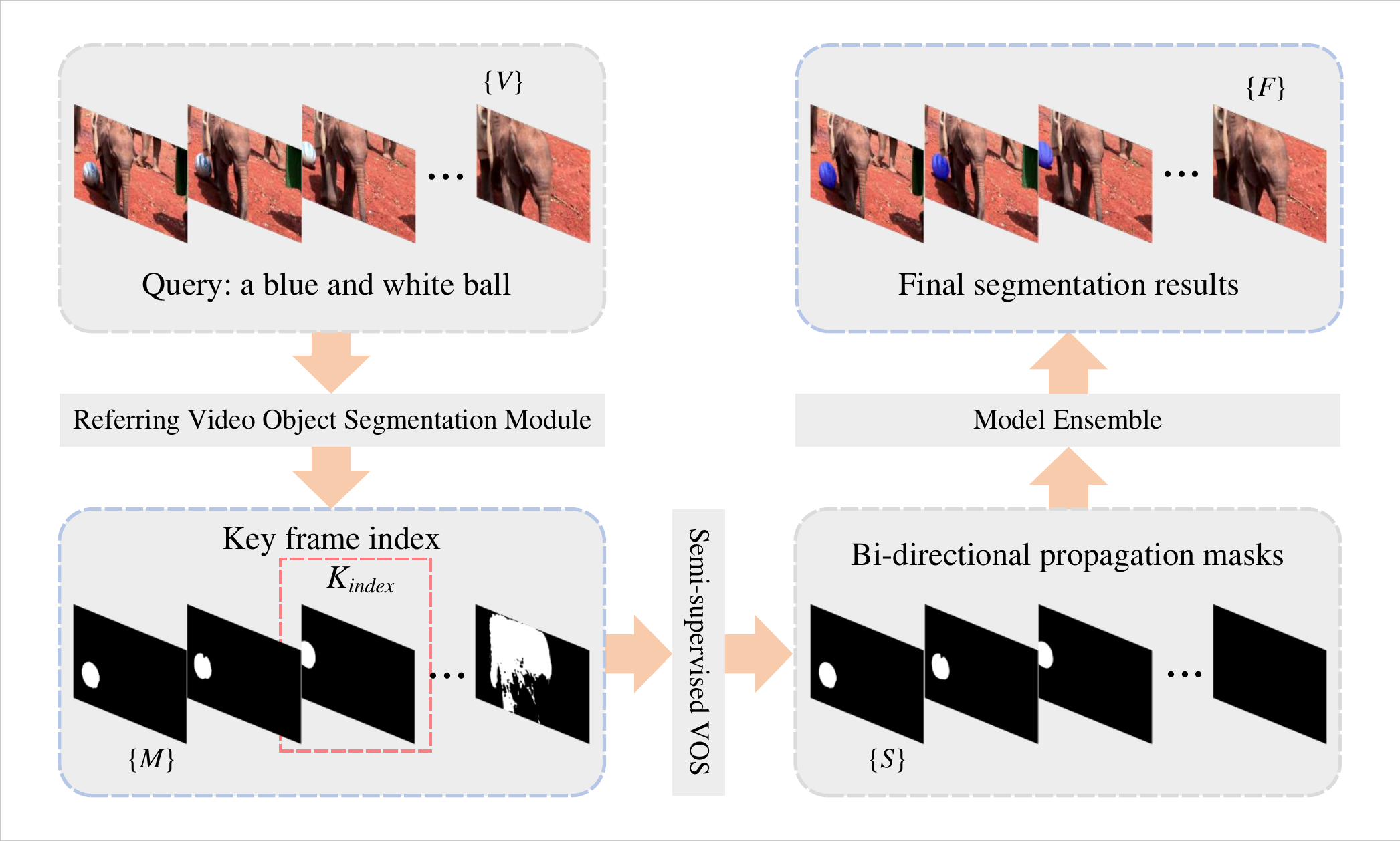}
\end{center}
\caption{The overall architecture of our model.}
\label{fig:fig1}
\end{figure}

In this work, we try to combine the advantages of the existing methods, to provide a simple and effective pipeline for RVOS task. We find that the multi-stage approach is mainly limited by the quality of the initial masks fed into the propagation model, and with high-quality masks related to the reference language can significantly improve the overall performance. While ReferFormer can provide masks that are strongly correlated with reference descriptions, but has certain limitations in temporal consistency. Based on the above observations, We first conduct extensive experiments on ReferFormer and improve the best single model a large margin on the validation set. Secondly, thanks to the high-quality mask sequences generated by ReferFormer, we demonstrate that the performance of the model can be further improved by a strong semi-supervised model based on simple keyframe selection.

Our method ranks 1st place in the 4th Large-scale Video Object Segmentation Challenge (CVPR2022): Referring Video Object Segmentation track~\cite{cvpr2022challenge}, with an overall $\mathcal{J} \& \mathcal{F}$ of 64.1 test-challenge.

\section{Related Work}
\label{sec:relate}

{\bf Semi-supervised Video Object Segmentation} The goal of semi-supervised VOS is to obtain pixel-level segmentation of objects across a video clip based on the mask annotation given at the first frame. The current mainstream methods~\cite{cheng2021rethinking, oh2019video, yang2020CFBI, yang2021aot} segment and track the target by matching the feature correlation between the target and the potential objects in the video sequence. STM~\cite{oh2019video} uses memory network to store object features from past frames and computes feature correlations based on attention mechanism. CFBI~\cite{yang2020CFBI} further enhances the accuracy of the correlation calculation by considering both foreground and background object features. To realize better and more efficient embedding learning, AOT~\cite{yang2021aot} employs an identification mechanism to associate multiple targets and a Long Short-Term Transformer to construct hierarchical matching and propagation. In our work, we utilize AOT for post-processing to improve the quality and temporal consistency of segmentations.

{\bf Referring Video Object Segmentation.} 
The RVOS task was first proposed by Gavrilyuk \textit{et al}~\cite{gavrilyuk2018actor}, whose goal is to segment and track actors and their actions in video content through natural language descriptions. The current method can be divided into two categories. (1) Multi-stage method. These methods~\cite{khoreva2018video, seo2020urvos, bellver2020refvos, liang2021rethinking} process each frame of the video clip separately through an image-level model. Representative works include URVOS~\cite{seo2020urvos}, which first performs initial mask prediction through an image-level model, and then propagates through a semi-supervised VOS method. (2) One-stage method. Recently, inspired by DETR~\cite{carion2020end}, ReferFormer~\cite{wu2022referformer} views the language as queries and directly attends to the most relevant regions in the video frames resulting in state-of-the-art performance. Our work draws on the advantages of the above two methods, obtains mask sequences strongly correlated with natural language descriptions based on ReferFormer, and further generates higher-quality results with the help of semi-supervised methods by selecting keyframes.

\begin{figure*}[t]
\begin{center}
\includegraphics[width=175mm]{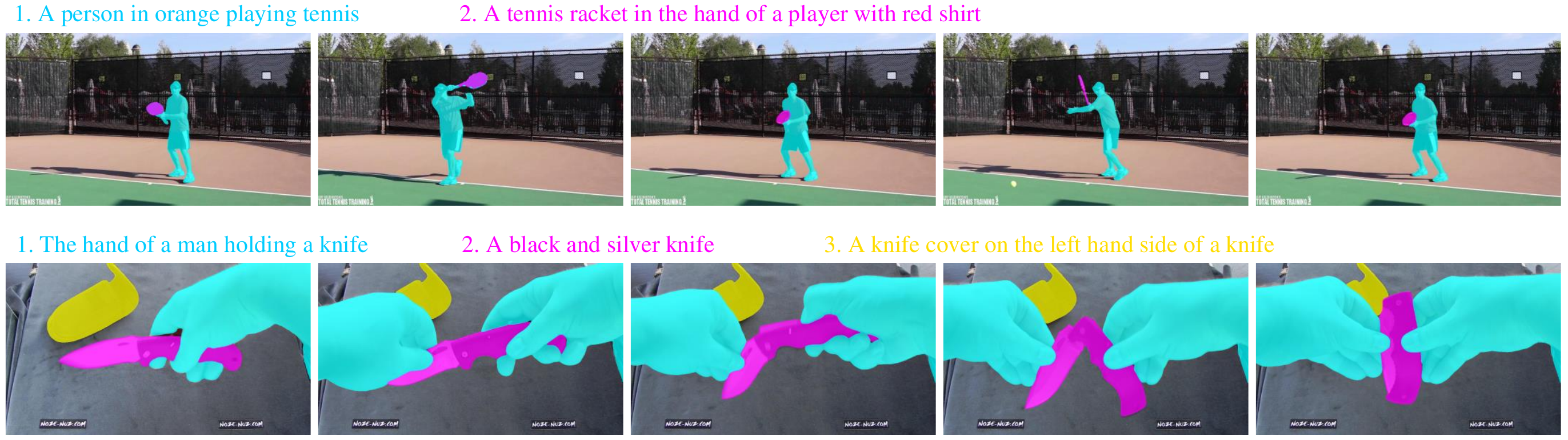}

\end{center}
\caption{Visualization results on Ref-Youtube-VOS.}
\label{fig:fig2}
\end{figure*}

\section{Method}

The input of RVOS contains a video sequence $\mathcal{V} = \left\{v_t\in \mathbb{R}^{C \times H \times W} \right\}_{t=1}^T $ with \textit{T} frames and a corresponding referring expression $\mathcal{E} = \left\{ e_l \right\}_{l=1}^L $ with \textit{L} words. We use the ReferFormer, a strong baseline of RVOS task, to obtain \textit{T}-frame binary segmentation masks $\mathcal{M} = \left\{ m_t \in \mathbb{R} ^ {H \times W} \right\}_{t=1}^T $. To further improve the quality and temporal consistency of the segmentation masks from the ReferFormer, we utilize the AOT algorithm to post-process our results. For the AOT post-process, we choose the frame with the highest score as the key-frame, and then use AOT to propagate it forward and backward to the entire video frames, producing high-quality results $\mathcal{S} = \left\{ s_t \in \mathbb{R} ^ {H \times W} \right\}_{t=1}^T $. Finally, we ensemble the results of multiple models of AOT 
to obtain the final segmentation masks $\mathcal{F} = \left\{ f_t \in \mathbb{R} ^ {H \times W} \right\}_{t=1}^T $. The overall architecture of the proposed method is illustrated in Figure~\ref{fig:fig1}. 

{\bf Backbone} As illustrated in Figure~\ref{fig:fig1}, the input of our framework consists of a video sequence $\mathcal{V}$ and a referring expression $\mathcal{E}$. We simply employ a universal RVOS framework as our backbone, \textit{i.e}., ReferFormer, which produces \textit{T}-frame binary segmentation masks $\mathcal{M}$ of referred object:
\begin{equation}
\mathcal{M} = \left\{\mathcal{F}^{ref}(\mathcal{E}, v_t)\right\}_{t=1}^T
\end{equation}
where $\mathcal{F}^{ref}$ denotes the ReferFormer model. During training, inspired by\cite{lin2021video}, we first use the image dataset RefCOCO and the video dataset Ref-Youtube-VOS to train the ReferFormer jointly, and then fine-tune it on the Ref-Youtube-VOS. In ~\cite{wu2022referformer}, the joint training process freezes the text encoder all the time, which may limit the guiding role of language. During the fine-tuning stage, we train the text encoder together with other modules to improve the language modeling ability.


{\bf Post-process} Previous work~\cite{wu2022referformer} has shown that using a semi-supervised VOS algorithm can further improve the accuracy of segmentation results and as model performance becomes stronger, the benefits of post-processing decrease. Our experiments find that even high-performing models can still achieve large gains when using a powerful semi-supervised VOS method. Given the ground-truth object masks of the first frame, semi-supervised VOS methods propagate the manual labeling to the entire video sequence. However, if we directly apply a semi-supervised VOS model to process our segmentation masks $\mathcal{M}$, some problems will occur. The object referred to by $\mathcal{E}$ may not appear in the first frame, and the quality of the segmentation results in the first frame may not be the best in the entire video sequence. Therefore, we need to seek a reasonable indicator to assist us select the frame with the highest segmentation quality in $\mathcal{M}$ as the key-frame for post-process. 

For the \textit{k}-th frame, the ReferFormer predicts the corresponding probability scalar $p_k \in \mathbb{R}^1 $ to indicate whether the prediction instance of the current frame corresponds to the referred object and the object is visible in the current frame. We first pick our key-frame index $\mathcal{K}_{index}$ using the probability scalar of the entire video sequence $\mathcal{P}$ :
\begin{equation}
\mathcal{K}_{index} = \arg\max(\mathcal{P})
\end{equation}
where $\mathcal{P} = \left\{ p_k \in \mathbb{R}^1 \right\}_{k=1}^T$. Then, we employ AOT to forward and backward propagate the key mask selected by key-frame index to the entire video clip and obtain corresponding object segmentation masks $\mathcal{S}$:
\begin{equation}
\mathcal{S} = \mathcal{F}^{post}(\mathcal{M}, \mathcal{K}_{index}) 
\end{equation}
where $\mathcal{F}^{post}$ denotes the AOT model.

{\bf Multi-model Fusion Based on Language Priors} By analyzing the Ref-Youtube-VOS dataset and final prediction results, we find that the same model predicts inconsistently when guided by different referring expressions with the same meaning. Similarly, different models predict inconsistently when guided by the same referring expression. To solve this problem, we fuse the masks predict by different referring expressions that describe the same target from different models. The fusion masks are voted at the pixel level. When the pixel value is greater than a certain threshold \textit{thr}, we divide the pixel into the foreground, otherwise, it is divided into the background.
\begin{equation}
y_t = \sum_{n=1}^N (s_t^n) 
\end{equation}
\begin{equation}
f_t^i=\begin{cases} 
0 & y_t^i<thr \\ 
1 & y_t^i>=thr
\end{cases}
\end{equation}
where $i \in \left\{ 1,2,...,HW \right\}$, \textit{N} denotes the number of results generated by different referring expressions with the same meaning in all models, $y_t \in \mathbb{R} ^ {H \times W} $ denotes the fusion mask of \textit{N} results. For the case that there is only one language description, we also fuse the results of all models and determine the final result according to the corresponding threshold value $thr_s$.


\begin{table}
\begin{center}
\begin{tabular}{c||c}
\toprule[1.1pt]
\textbf{Model} & {$\mathcal{J}$ \& $\mathcal{F}$} $\uparrow$ \\ 
\hline\hline
Baseline & 64.9 \\
\hline
+Finetune on Ref-Youtube-VOS dataset & 66.0 \textcolor{red}{(+1.1)} \\
+Key-frame \& AOT & 70.3 \textcolor{red}{(+4.3)}\\
+Multi-model Fusion \& AOT & 71.0 \textcolor{red}{(+0.7)} \\
+Model Ensemble & 72.4 \textcolor{red}{(+1.4)} \\
\hline
\end{tabular}
\end{center}
\caption{Ablation study of each module on our model’s performance on \textbf{validation set}.}
\label{table:two}
\end{table}

\section{Experiment}

{\bf Dataset and Metrics} We measure the effectiveness of our model on \textit{2022 Referring Youtube-VOS challenge}~\cite{cvpr2022challenge}, which is based on YouTube-VOS-2019 dataset~\cite{seo2020urvos}. Ref-Youtube-VOS dataset has 3,978 high-resolution YouTube videos with about 15K language expressions. These video are divided into 3,471 training videos, 202 validation videos and 305 test videos. We use two standard metrics, \textit{i.e.}, region similarity $\mathcal{J}$ and contour accuracy $\mathcal{F}$ following~\cite{perazzi2016benchmark}, for evaluation.

{\bf Detailed Network Architecture} We employ two simple and powerful benchmark networks, ReferFormer and AOT. For ReferFormer, we adopt Video-Swin-Base~\cite{liu2021video} as the visual encoder and RoBERTa-Base~\cite{liu2019roberta} as the text encoder. For the mask propagation model AOT, we adopt Swin-L~\cite{liu2021swin} as the backbone. 

{\bf Training Detail.} During fine-tuning, ReferFormer is trained on Ref-Youtube-VOS dataset, optimized using AdamW optimizer with the weight decay of 5e-4, a learning rate of 5e-6, and an initial learning rate of 1e-6 for the rest. We fine-tune the model for 6 epochs with the learning rate decays divided by 10 at 3-th and 5-th epoch. It should be noted that we do not freeze text encoder parameters during fine-tuning. In the post-process stage, we retrain the AOT network with Swin-L as the backbone, and the specific training parameters are consistent with the default AOT~\cite{yang2021aot} setting.

{\bf Model Ensemble} To further improve the segmentation accuracy, we utilize the model ensemble strategy which is the same as multi-model fusion based on language priors to fuse the results of the fine-tuning model, the ReferFormer official Video-Swin-Base model and the Swin-L model, and send the fusion masks to the AOT for post-process, the key-frame index is derived from the fine-tuning model. Finally, we ensemble the AOT results of the fusion model, fine-tuning model, ReferFormer official Video-Swin-Base model and Swin-L model to get the final submission masks.

{\bf Results on RVOS Challenge.} Our approach achieves 64.13  on the final leaderboard, ranking 1st place on CVPR2022 Referring Youtube-VOS challenge and outperforming the next best team by 2.4\% in the aspect of overall $\mathcal{J} \& \mathcal{F}$.

{\bf Ablation Study.} To study the effect of each module on our model’s performance, we start our ablation study with a simple baseline network, \textit{i.e.}, ReferFormer, as illustrated in Table~\ref{table:two}. We first finetune our model on Ref-Youtube-VOS dataset, and it improves performance by 1.1\%. A reasonable key-frame selection strategy combined with AOT post-process can achieve significant performance improvement ($3^{\textit{rd}}$ row in Table~\ref{table:two}). Then, we utilize the model ensemble strategy to fuse the masks of the fine-tuning model, the ReferFormer official Video-Swin-Base model and the Swin-L model, and send the fusion masks to the AOT for further post-process, and it brings a 0.7\% performance boost. Finally, again using the model ensemble scheme to fuse the AOT results of multi-model, the fine-tuning model, the ReferFormer official Video-Swin-Base model and the Swin-L model, we achieve a performance of 72.4\% on the validation set.

{\small
\bibliographystyle{ieee_fullname}
\bibliography{egbib}

\begin{thebibliography}{10}\itemsep=-1pt

\bibitem{cvpr2022challenge}
The 4th large-scale video object segmentation challenge.
\newblock \url{https://youtube-vos.org/challenge/2022}.

\bibitem{bellver2020refvos}
Miriam Bellver, Carles Ventura, Carina Silberer, Ioannis Kazakos, Jordi Torres,
  and Xavier Giro-i Nieto.
\newblock Refvos: A closer look at referring expressions for video object
  segmentation.
\newblock {\em arXiv preprint arXiv:2010.00263}, 2020.

\bibitem{carion2020end}
Nicolas Carion, Francisco Massa, Gabriel Synnaeve, Nicolas Usunier, Alexander
  Kirillov, and Sergey Zagoruyko.
\newblock End-to-end object detection with transformers.
\newblock In {\em European conference on computer vision}, pages 213--229.
  Springer, 2020.

\bibitem{cheng2021rethinking}
Ho~Kei Cheng, Yu-Wing Tai, and Chi-Keung Tang.
\newblock Rethinking space-time networks with improved memory coverage for
  efficient video object segmentation.
\newblock {\em Advances in Neural Information Processing Systems}, 34, 2021.

\bibitem{gavrilyuk2018actor}
Kirill Gavrilyuk, Amir Ghodrati, Zhenyang Li, and Cees~GM Snoek.
\newblock Actor and action video segmentation from a sentence.
\newblock In {\em Proceedings of the IEEE Conference on Computer Vision and
  Pattern Recognition}, pages 5958--5966, 2018.

\bibitem{khoreva2018video}
Anna Khoreva, Anna Rohrbach, and Bernt Schiele.
\newblock Video object segmentation with language referring expressions.
\newblock In {\em Asian Conference on Computer Vision}, pages 123--141.
  Springer, 2018.

\bibitem{liang2021rethinking}
Chen Liang, Yu Wu, Tianfei Zhou, Wenguan Wang, Zongxin Yang, Yunchao Wei, and
  Yi Yang.
\newblock Rethinking cross-modal interaction from a top-down perspective for
  referring video object segmentation.
\newblock {\em arXiv preprint arXiv:2106.01061}, 2021.

\bibitem{lin2021video}
Huaijia Lin, Ruizheng Wu, Shu Liu, Jiangbo Lu, and Jiaya Jia.
\newblock Video instance segmentation with a propose-reduce paradigm.
\newblock In {\em Proceedings of the IEEE/CVF International Conference on
  Computer Vision}, pages 1739--1748, 2021.

\bibitem{liu2019roberta}
Yinhan Liu, Myle Ott, Naman Goyal, Jingfei Du, Mandar Joshi, Danqi Chen, Omer
  Levy, Mike Lewis, Luke Zettlemoyer, and Veselin Stoyanov.
\newblock Roberta: A robustly optimized bert pretraining approach.
\newblock {\em arXiv preprint arXiv:1907.11692}, 2019.

\bibitem{liu2021swin}
Ze Liu, Yutong Lin, Yue Cao, Han Hu, Yixuan Wei, Zheng Zhang, Stephen Lin, and
  Baining Guo.
\newblock Swin transformer: Hierarchical vision transformer using shifted
  windows.
\newblock In {\em Proceedings of the IEEE/CVF International Conference on
  Computer Vision}, pages 10012--10022, 2021.

\bibitem{liu2021video}
Ze Liu, Jia Ning, Yue Cao, Yixuan Wei, Zheng Zhang, Stephen Lin, and Han Hu.
\newblock Video swin transformer.
\newblock {\em arXiv preprint arXiv:2106.13230}, 2021.

\bibitem{oh2019video}
Seoung~Wug Oh, Joon-Young Lee, Ning Xu, and Seon~Joo Kim.
\newblock Video object segmentation using space-time memory networks.
\newblock In {\em Proceedings of the IEEE/CVF International Conference on
  Computer Vision}, pages 9226--9235, 2019.

\bibitem{perazzi2016benchmark}
Federico Perazzi, Jordi Pont-Tuset, Brian McWilliams, Luc Van~Gool, Markus
  Gross, and Alexander Sorkine-Hornung.
\newblock A benchmark dataset and evaluation methodology for video object
  segmentation.
\newblock In {\em Proceedings of the IEEE conference on computer vision and
  pattern recognition}, pages 724--732, 2016.

\bibitem{seo2020urvos}
Seonguk Seo, Joon-Young Lee, and Bohyung Han.
\newblock Urvos: Unified referring video object segmentation network with a
  large-scale benchmark.
\newblock In {\em European Conference on Computer Vision}, pages 208--223.
  Springer, 2020.

\bibitem{vaswani2017attention}
Ashish Vaswani, Noam Shazeer, Niki Parmar, Jakob Uszkoreit, Llion Jones,
  Aidan~N Gomez, {\L}ukasz Kaiser, and Illia Polosukhin.
\newblock Attention is all you need.
\newblock {\em Advances in neural information processing systems}, 30, 2017.

\bibitem{wu2022referformer}
Jiannan Wu, Yi Jiang, Peize Sun, Zehuan Yuan, and Ping Luo.
\newblock Language as queries for referring video object segmentation.
\newblock {\em arXiv preprint arXiv:2201.00487}, 2022.

\bibitem{yang2020CFBI}
Zongxin Yang, Yunchao Wei, and Yi Yang.
\newblock Collaborative video object segmentation by foreground-background
  integration.
\newblock In {\em European Conference on Computer Vision}, pages 332--348.
  Springer, 2020.

\bibitem{yang2021aot}
Zongxin Yang, Yunchao Wei, and Yi Yang.
\newblock Associating objects with transformers for video object segmentation.
\newblock In {\em Advances in Neural Information Processing Systems (NeurIPS)},
  2021.

\end{thebibliography}
}

\end{document}